\documentclass[9pt,shortpaper,twoside,web]{ieeecolor}
\usepackage{generic}
\usepackage{amsmath,amssymb,amsfonts}
\usepackage{algorithmic}
\usepackage{graphicx}
\usepackage{textcomp}

\usepackage[style=ieee,citestyle=ieee]{biblatex}
\bibliography{references}

\usepackage{color, colortbl} 
\definecolor{Gray}{gray}{0.9}

\def\BibTeX{{\rm B\kern-.05em{\sc i\kern-.025em b}\kern-.08em
    T\kern-.1667em\lower.7ex\hbox{E}\kern-.125emX}}
\begin{document}

\title{Automated Respiratory Event Detection Using Deep Neural Networks}
\author{Thijs E Nassi, Wolfgang Ganglberger, Haoqi Sun, Abigail A Bucklin, Siddharth Biswal, Michel J A M van Putten, Robert J Thomas, M Brandon Westover

\thanks{M.B.W. Was supported by the Glenn Foundation for Medical Research and American Federation for Aging Research (Breakthroughs in Gerontology Grant); American Academy of Sleep Medicine (AASM Foundation Strategic Research Award); Football Players Health Study (FPHS) at Harvard University; Department of Defense through a subcontract from Moberg ICU Solutions, Inc; by the NIH (1R01NS102190, 1R01NS102574, 1R01NS107291, 1RF1AG064312). (Corresponding author: M. Brandon Westover.)}
\thanks{T.E. Nassi and M.J.A.M Van Putten are with University of Twente, 7522NB Enschede, the Netherlands (e-mail: t.nassi@student.utwente.nl; m.j.a.m.vanputten@utwente.nl).}
\thanks{W. Ganglberger, H. Sun, A.A. Bucklin and M.B. Westover are with Massachusetts General Hospital, Boston, MA, 02114 USA (e-mail: wganglberger@mgh.harvard.edu; abucklin@partners.org; mwestover@mgh.harvard.edu). }
\thanks{S. Biswal is with School of Computational Science and Engineering, Georgia Institute of Technology, Atlanta, GA, USA (e-mail: siddnitr1@gmail.com). }
\thanks{R.J. Thomas is with Deaconess Medical Center, Boston, MA, 02215, USA (e-mail: rthomas1@bidmc.harvard.edu).}}

\maketitle

\begin{abstract}
The gold standard to assess respiration during sleep is polysomnography; a technique that is burdensome, expensive (both in analysis time and measurement costs), and difficult to repeat. Automation of respiratory analysis can improve test efficiency and enable accessible implementation opportunities worldwide. 
Using 9,656 polysomnography recordings from the Massachusetts General Hospital (MGH), we trained a neural network (WaveNet) based on a single respiratory effort belt to detect obstructive apnea, central apnea, hypopnea and respiratory-effort related arousals. Performance evaluation included event-based and recording-based metrics – using an apnea-hypopnea index analysis. The model was further evaluated on a public dataset, the Sleep-Heart-Health-Study-1, containing 8,455 polysomnographic recordings.
For binary apnea event detection in the MGH dataset, the neural network obtained an accuracy of 95\%, an apnea-hypopnea index r\textsuperscript{2} of 0.89 and area under the curve for the receiver operating characteristics curve and precision-recall curve of 0.93 and 0.74, respectively. For the multiclass task, we obtained varying performances: 81\% of all labeled central apneas were correctly classified, whereas this metric was 46\% for obstructive apneas, 29\% for respiratory effort related arousals and 16\% for hypopneas. The majority of false predictions were misclassifications as another type of respiratory event.
Our fully automated method can detect respiratory events and assess the apnea-hypopnea index with sufficient accuracy for clinical utilization. Differentiation of event types is more difficult and may reflect in part the complexity of human respiratory output and some degree of arbitrariness in the clinical thresholds and criteria used during manual annotation. 
\end{abstract}


\section{Introduction}
\label{sec:introduction}
Sleep disorders such as sleep apnea and insomnia affect millions of people worldwide \cite{Benjafield2019}. Clinical effects include difficulty in initiating and maintaining sleep, impaired alertness, and hypertension. Excessive daytime sleepiness and fatigue, two common symptoms associated with sleep disorders, have a large impact on population health \cite{Skaer2010, Pietzsch2011}. Accurate and timely diagnosis of a patient’s sleep disorder is therefore essential. Patients with apnea, especially obstructive sleep apnea, are at increased risk for traffic accidents, postoperative complications, and delirium \cite{Avidan2018,Revels2019}. Untreated sleep apnea is associated with arrhythmias, heart failure and stroke. Studies that measure the apnea-hypopnea index (AHI) show that an estimated 49.7\% of male and 23.4\% of female adults have moderate-to-severe sleep-disordered breathing, though a lower percentage are clinically symptomatic \cite{Revels2019}. 

The gold standard to measure sleep objectively is laboratory-based polysomnography (PSG). PSG is conventionally scored based on the American Academy of Sleep Medicine (AASM) guidelines. Scoring PSG recordings is a time-consuming task performed by specialists in dedicated sleep centers, making this an expensive process both in time and costs. Automation of PSG analysis would decrease the required analysis time and reduce costs. Moreover, automated PSG analysis computer models could be implemented in clinical centers anywhere in the world and across a variety of data acquisition options, including home sleep testing, testing in acute care environments, specific operational conditions such as high altitude, and consumer wearable devices.

Medical data is complex and involves a large number of variables and context that are difficult to encompass by programs based on a fixed set of rules. Deep learning models such as convolutional neural networks (CNN) and recurrent neural networks (RNN) have been applied in many domains to solve complex pattern recognition tasks \cite{Najafabadi2015}. Deep learning algorithms rely on patterns and inference rather than explicit instructions and can learn intricate relationships between features and labels from data. Implementing neural networks has become relevant in analyzing the heterogeneous kinds of data generated in modern clinical care \cite{Yue2020}. Various types of deep learning algorithms have been found to be suitable for analyzing specific types of data. For instance, CNNs have been successful in classifying objects in images. Typical CNN architectures, however, are not ideal when analyzing temporal data. Temporal data is typically better exploited by RNNs. However, the recently introduced CNN, WaveNet architecture has been found to perform better than RNNs on several tasks \cite{Oord2016}. WaveNet's architecture resembles a typical CNN, yet the application of dilated causal convolutions creates an effectively larger receptive field. This renders WaveNet capable of detecting both spatial patterns and long-range temporal patterns. WaveNet was originally designed to synthesize speech; however, its application has been found suitable for analyzing other types of signals. In 2018 a challenge organized by the PhysioNet Computing in Cardiology aimed to detect sleep arousals from a variety of physiological signals, including signals derived from respiration. The winning model was a modified WaveNet architecture, suggesting that this CNN architecture can indeed perform successfully in other domains such as the automation of PSG-related tasks \cite{Zabihi2019}.

In the last two years a significant number of papers have been published on the detection of sleep apnea, as described by recent review papers \cite{Mostafa2019,Uddin2018}. Finding a patient-friendly and accurate sensor or signal, especially in combination with a suitable analysis model, is clearly an ongoing area of high relevance. An overview of other sleep apnea studies can be found in Table \ref{tab1}. 

\begin{table*}
\small
\centering
\caption{Overview of other studies performing automated respiratory event detection}
\begin{tabular}{|c c l c c | c c c c c|}
    \hline
    Study & Dataset & Signal type & Analysis & Classifier & Accuracy & Sensitivity & Specificity & Precision & F1-score \\
    & size & & model & type & \footnotesize{(\%)} & \footnotesize{(\%)} & \footnotesize{(\%)} & \footnotesize{(\%)} & \footnotesize{(\%)} \\
    \hline
    \cite{Biswal2018}               & 10.000    & Airflow, respiration chest and    & RCNN      & G     & 88.2  & -     & -     & -     & - \\ 
    && abdomen, oxygen saturation &&&&&&& \\ 
    \cite{VanSteenkiste2019}        & 2100      & Respiration abdomen               & LSTM      & A/N   & 77.2  & 62.3  & 80.3  & 39.9  & - \\ 
    \cite{Haidar2018}               & 1507      & Nasal airflow, abdominal and      & CNN1D-3ch &       & 83.5  & 83.4  & -     & 83.4  & 83.4  \\
    &&  thoracic plethysmography &&&&&&& \\
    \cite{McCloskey2018}            & 1507      & Nasal airflow             & CNN2D         & A/H/N & 79.8  & 79.9  & -     & 79.8  & 79.7  \\   
    \cite{Banluesombatkul2018}      & 545       & Electrocardiography       & CNN1D-LSTM   & G     & 79.5  & 77.6  & 80.1  & -     & 79.1  \\
    &&& MHLNN &&&&&& \\   
    \cite{Lakhan2018}               & 520       & Airflow                           & MHLNN     & G         & 87.2      & 88.3     & 87.8   & -     & - \\   
    \cite{Espinoza-Cuadros2015}     & 285       & Voice and facial features         & GMM       & G         & 72        & 73       & 65     & -     & - \\ 
    \cite{Gutierrez-Tobal2015}      & 188       & Airflow, respiratory rate         & LR        & G         & 72        & 80       & 59     & -     & - \\ 
    && variability   &&&&&&& \\   
    \cite{Alvarez2006}      & 187       & pulse oximetry       & CTM       & G      & 87       & 90         & 83        & -         & - \\   
    \cite{Rosenwein2015}    & 186       & Breathing sounds     & Binary-RF & G      & 86       & -          & -         & -         & - \\   
    \cite{Choi2018}         & 179       & Nasal pressure       & CNN1D     & A/H/N  & 96.6     & 81.1       & 98.5      & 87        & - \\   
    \cite{Kim2018}          & 120       & Breathing sounds     & MHLNN     & G      & 75       & -          & -         & -         & - \\   
    \cite{Haidar2017}       & 100       & Nasal airflow        & CNN1D     & OA/N   & 74.7     & 74.7       & -         & 74.5      & - \\   \hline
\multicolumn{10}{p{18.1cm}}{Analysis models: RCNN $=$ recurrent and convolutional neural networks, LSTM $=$ long short-term memory, CNN $=$ convolution neural network, MHLNN $=$ multiple hidden layers neural network, GMM $=$ gaussian mixture model, LR $=$ logistic regression, CTM $=$ central tendency measure, RF $=$ random forest. Classifier types: A $=$ apnea, H $=$ hypopnea, N $=$ normal, O $=$ obstructive, G $=$ global. }
\end{tabular}
\label{tab1}
\end{table*} 

Sleep apnea detection methods typically use various breathing measurements and oximetry \cite{Uddin2018}. Alternative methods using signals derived from electrocardiography (ECG) have shown some promise for predicting AHI as well, although such data has an indirect relationship to the respiratory system and therefore to sleep apnea \cite{Bsoul2011,Mostafa2019}. This more indirect method of analyzing respiration requires additional processing and can be affected by other illnesses including heart failure and cardiac arrhythmias, rather than sleep apnea \cite{VanSteenkiste2019}. Using multiple physiological signals to detect sleep apnea can provide good performance \cite{Haidar2017,Biswal2018}. However, this leads to similar problems as the current gold standard; using many different sensor signals is considered uncomfortable, expensive, and time-consuming. An autonomous model that achieves acceptable performance while using a more patient friendly approach than the current gold standard is an unmet need in existing literature, however it is important for large-scale clinical implementation. The ability to identify and discriminate between the specific respiratory events that are typically scored in PSG while using fewer signals is also unknown to the current clinical setting. In this research we aimed to create a fully automated method that can detect respiratory events, discriminate between the different types of respiratory events, and assess the apnea-hypopnea index with sufficient efficiency for clinical implementation using only a single respiratory effort belt.

\section{Methods}
\subsection{Dataset}
The dataset used to train our model was from The Massachusetts General Hospital (MGH) sleep laboratory. The MGH Institutional Review Board approved the retrospective analysis of the clinically acquired PSG data. In total 9656 PSG recordings were successfully exported and split into training, validation, and testing subsets. To obtain a representative and heterogeneous test set we selected 1000 patients with the following approach. For the four categories sex, age, body mass index (BMI) and AHI we stratified the dataset into subgroups, see Table \ref{tab2}. We randomly selected 250 patients in each category to ensure equal numbers of patients in each subgroup. All other data was used for training (7856 patients) and validation (800 patients). There was no overlap in patients between training and test sets. Patients with and without breathing assistance by continuous positive airway pressure (CPAP) were included. 

We included a secondary dataset for external validation of our model. This dataset was collected by the Sleep Heart Health Study (SHHS) and included 8455 PSG recordings. For this research we only used the signal measured at the abdomen using a respiratory effort belt (inductance plethysmography). This signal, in comparison to the available respiratory signals measured on the thorax, is expected to provide the best predictive performance \cite{VanSteenkiste2018}. 

All recordings from the MGH were annotated by sleep experts following AASM guidelines. Specifically, respiratory event detections included obstructive apneas, central apneas, mixed apneas, hypopneas, and respiratory effort-related arousals (RERA’s). We define respiratory events as any of those four events and apnea events as any type of apnea or hypopnea

Recordings obtained from the SHHS database were annotated according to SHHS guidelines. A key difference between the two datasets is the primary respiratory scoring signal in the original source – nasal pressure (MGH) and thermistor (SHHS). This difference and implications will be discussed further below.

\begin{table}
\centering
\smallskip
\caption{Dataset (N=9656) distribution containing four categories and their associated patient bins}
\setlength{\tabcolsep}{3pt}
\begin{tabular}{|rlc|}
\hline
Category & Bin & Percentage of all patients \\

\hline
        Sex & male & 58.9\% \\
        & female & 40.7\% \\
        & unknown & 0.4\% \\
\hline
        Age & \footnotesize{$<$} \normalsize 60 & 65.0\%\\
        & 60 - 80 & 34.4\%\\
        & \footnotesize{$>$} \normalsize 80 & 2.6\% \\
\hline
        BMI & underweight \footnotesize{($<$ 18.5)} & 0.9\% \\
        & normal weight \footnotesize{(18.5 - 25)} & 13.7\% \\
        & overweight \footnotesize{(25 - 30)} & 26.7\%\\
        & obese \footnotesize{($>$ 30)} & 58.7\%\\
\hline
        AHI & normal \footnotesize{($<$ 5)} & 39.9\% \\
        & mild \footnotesize{(5 - 15)} & 27.7\%  \\
        & moderate \footnotesize{(15 - 30)}  & 20.4\% \\
        & severe \footnotesize{($>$ 30)} & 12.0\% \\
\hline
\end{tabular}
\label{tab2}
\end{table}

\subsection{Preprocessing and data preparation}
All recordings that were incomplete or did not include any sleep were removed. Next, to extract the relevant respiratory information and remove present noise, minimal preprocessing techniques were applied. The abdominal respiration measurement from both the MGH data and the SHHS data consisted of a single channel with a sampling frequency of 125 Hz, 200 Hz or 250 Hz. All recordings were resampled to 10 Hz and a notch filter of 60 Hz was applied to reduce line noise. A low-pass filter of 10 Hz was applied to remove higher frequencies not of interest. Z-score normalization was performed using the mean and standard deviation of the 1\textsuperscript{st} to 99\textsuperscript{th} percentile clipped signal to optimize the training process of the neural network.

The training data was segmented into 7-minute epochs with a stride of 30 seconds. Each epoch was assigned one ground truth class label – the sleep expert’s label located in the center of the segment. We segmented the test data in the same way, except that we use a stride of 1 second, which allowed for a respiratory event prediction for each second.

\subsection{Model and prediction tasks}
In this research we utilized a deep neural network (‘WaveNet’, see model architecture in Section \ref{sec:arch}) to automatically detect apneas and hypopneas from a single effort belt signal, without use of additional sensors that are conventional in PSG measurements (e.g. thermistor, nasal pressure, oxygen saturation, electroencephalography or electrocardiography), and without using human-engineered features. As described above, the signal was segmented into 7-minute epochs and, in this way, the model was trained to predict only the center index of a 7-minute epoch, while having 3.5 minutes of context information before and after the center index. We designed the following two prediction tasks:

\begin{itemize}
    \item Binary classification to discriminate non-apnea events from apnea-hypopnea events (regular breathing and respiratory events). Based on the predicted respiratory events, we computed the predicted AHI as the number of predicted respiratory events per hour of sleep. 
    \item Multiclass classification to discriminate the respiratory event classes: no-event, obstructive apnea or mixed apnea, central apnea, RERA, hypopnea. From the sum of the detected respiratory events, we determined the AHI and respiratory disturbance index (RDI).
\end{itemize}

\begin{figure*}[!t]
\centerline{\includegraphics[width=\textwidth]{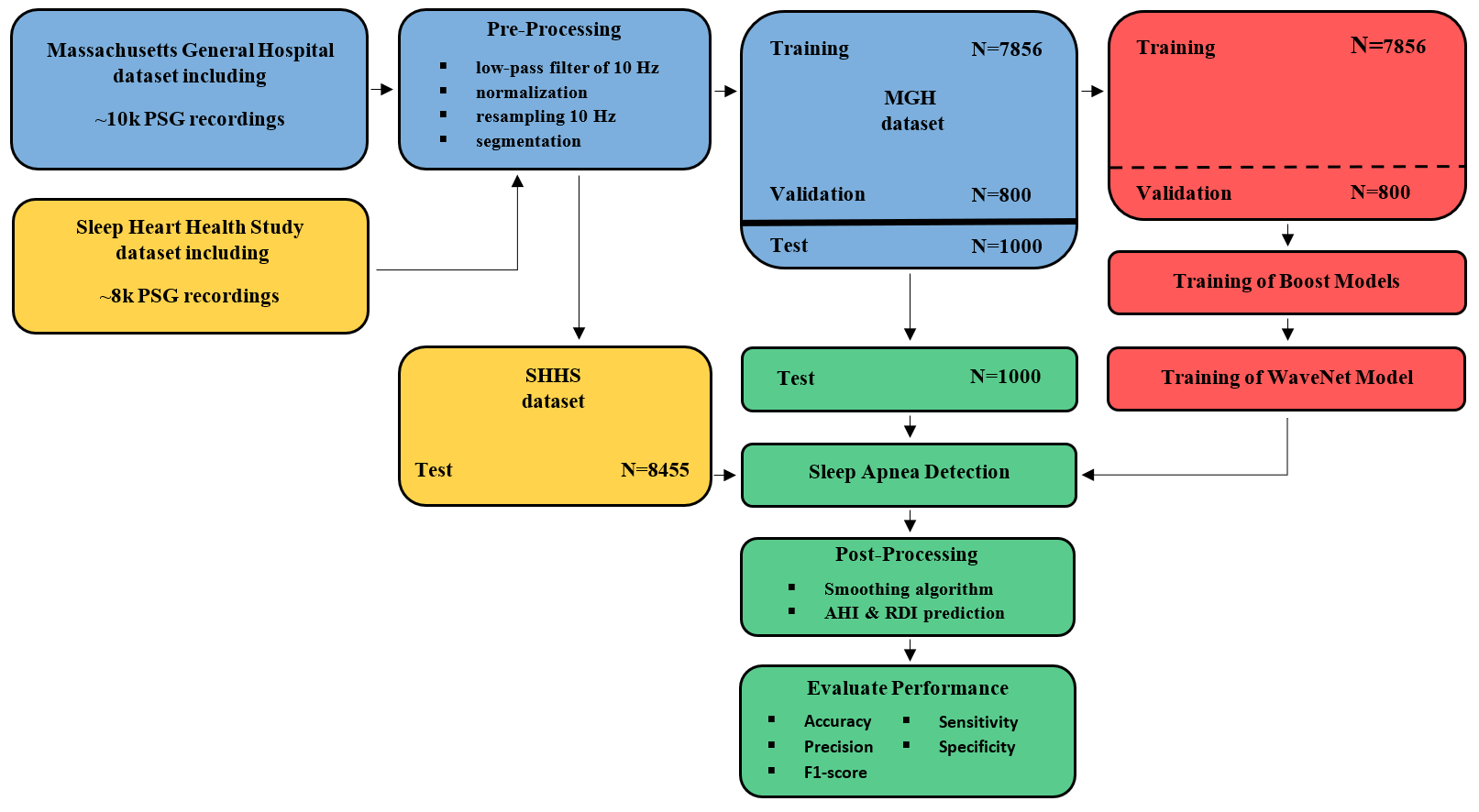}}
\centering
\caption{Data flow scheme for model development and testing. The model was trained and validated on the dataset from the Massachusetts General Hospital (MGH) whereas the dataset from the Sleep Heart Health Study (SHHS) was used for external validation. Both the Apnea Hypopnea Index (AHI) and the respiratory disturbance index (RDI) were computed during post-processing.}
\label{fig1}
\end{figure*}

In each experiment our model provided a probability for all included classes. The highest probability among the possible classes constitutes the output of our model. In Fig. \ref{fig1} the complete workflow scheme is shown.

\subsection{Model architecture}
\label{sec:arch}
We chose for our model architecture a WaveNet network \cite{Oord2016,GitWaveNet}. The use of dilated causal convolutions, which exhibit very large receptive fields, is useful for analyzing temporal data and therefore respiration. Exponentially increasing the dilation factor results in exponential receptive field growth with each hidden layer. Using 12 layers resulted in a receptive field of 4096 samples in our 10 Hz signal, representing approximately 7 minutes. A dropout of 0.2 was used to improve generalization of the network. The number of filters for each of the convolutions was set to 32. A kernel size of 2 was used. The categorical cross entropy loss function was applied during training, 

\begin{equation}
\mbox{loss} = \sum\limits_{i=1}^N -{y_i^\prime} log(y_i), \\
\end{equation}

where $y$ represents the predicted probability distribution, $y\prime$ represents the true distribution, and $N$ represents the number of classes.

\subsection{Boosting for imbalanced data}
Classification with imbalanced data is challenging in many real-world deep learning applications \cite{Buda2017,Johnson2019}. For the PSG recordings in our research, the number of epochs containing only regular breathing is typically much larger than the epochs containing respiratory events, even for patients classified with severe apnea. For this problem we designed a boosted model approach by applying a binary WaveNet classifier, or boost-model, over multiple iterations. To remove a large proportion of epochs with regular breathing without removing many epochs including apnea events, only epochs with an extremely high probability of regular breathing were removed by the boost-model. In our approach we selected a probability threshold to make our boost-model extremely sensitive for apneas, based on the receiver operating characteristic (ROC) curve. In the first iteration we used a TPR of 0.995 and decreased this value by 0.010 for each subsequent iteration. The boosted model iterations stopped when the desired balance in classes was obtained. This balance was defined by 3:1 ratio of regular breathing with respect to the sum of events in experiment 1, and a 3:1 ratio of regular breathing with respect to the most commonly occurring event type in experiment 2. 

In each iteration, the boost model received non-rejected samples from the previous iteration. Using this approach, the boost-model was trained to discriminate regular breathing from other respiratory events, while being exposed to a decreasing and increasingly more challenging dataset. In this way, in every iteration our boost-model should learn new nuanced characteristics that define a normal breathing rhythm. Using this boosted approach, we vastly reduced the number of epochs containing regular breathing and improved effective classification by our main WaveNet model. Moreover, the boost-model was expected to remove epochs with regular breathing that are relatively simple to distinguish from apnea while leaving the more complex epochs for our main model. All boost-models were defined by the same hyperparameters as our main model.

\subsection{Post-processing}
After applying our model, we obtained an apnea prediction for each second. The prediction resolution of 1 Hz allowed high fluctuations of predicted events and allowed detection of very brief events. Both situations were considered not physiologically plausible, therefore, we designed a smoothing algorithm. This smoothing algorithm removed short events and rapidly changing events. The algorithm was based on a moving window of 10 seconds. The following rules were applied for each window. 

\begin{itemize}
    \item When a minimum of 3 out of 10 seconds was classified as no-event, the complete window was set to no-event.
    \item If a window was classified as an event and multiple types of events were present, the type of event that occurred most became the prediction for the complete window. 
\end{itemize}

Finally, consecutive events with a combined length of two windows or greater were converted into a single event prediction. The type of event that held the largest proportion among the combined events indicated this new prediction. Applying the smoothing algorithm resulted in predicted events with a minimum length of 10 seconds, similar to what is suggested by AASM guidelines. 

To obtain a more clinically interpretable performance granularity, we specified the following criterion to judge when a detected event overlapped sufficiently with an event annotated by sleep experts to count as correct: 
\begin{itemize}
    \item A predicted event is considered a correct prediction when more than 50\% of its duration overlaps with an expert label.
\end{itemize}

\subsection{Model evaluation}
Confusion matrices were computed to assess per-event performance of our model. The true positives (TP), true negatives (TN), false positives (FP) and false negatives (FN) were computed for each of the 2 or 5 classes, respectively, for experiment 1 and 2. To determine the TN value, a quantization of similar granularity, i.e. per event, was calculated. This was accomplished by dividing the total duration of regular breathing by the median length of all respiratory events in the MGH test data, 18 seconds. Next, the TP, TN, FP and FN values were used to determine the following event-per-event performance metrics of our model. 
\bigskip

$
\begin{array}{lcl}
\mbox{Accuracy} & = & \frac{\mbox{TP} + \mbox{TN}} {\mbox{TP} + \mbox{TN} + \mbox{FP} + \mbox{FN}} \\\\
\mbox{Sensitivity} & = & \frac{\mbox{TP}} {\mbox{TP} + \mbox{FN}} \\\\
\mbox{Specificity} & = & \frac{\mbox{TN}} {\mbox{TN} + \mbox{FP}} \\\\
\mbox{Precision} & = & \frac{\mbox{TP}} {\mbox{TP} + \mbox{FP}} \\\\
\mbox{F1 score} & = & \frac{2 \times  \mbox{Sensitivity} \times  \mbox{Precision}} { \mbox{Sensitivity} +  \mbox{Precision} }
\end{array}
$
\bigskip

Additionally, for the binary tasks the ROC curve and precision-recall curve and their corresponding areas (AUC\textsubscript{ROC}  AUC\textsubscript{PRC}) were computed. 

In addition to event-per-event evaluation, we evaluated global scoring performance. Global assessment of sleep apnea severity is typically used for clinical diagnosis \cite{Mostafa2019}. For the first experiment we determined the AHI value per patient, whereas in the second experiment we determined the AHI and RDI value per patient using the following computations.
\bigskip

$
\begin{array}{lcl}
\mbox{AHI} & = & \frac{\mbox{OA} + \mbox{CA} + \mbox{HY}} {\mbox{hours of sleep }} \\
\\
\mbox{RDI} & = & \frac{\mbox{OA} + \mbox{CA} + \mbox{HY} + \mbox{RERA’s}} {\mbox{hours  of  sleep }} \\
\end{array}
$
\bigskip

where OA is obstructive apneas, CA is central apneas, and HY is hypopneas. With the AHI score all patients were categorized as normal or mild, moderate or severe sleep apnea. Categorization was according to conventional criteria as defined by AASM guidelines.
\begin{itemize}
    \item Normal breathing: AHI $\leq$ 5
    \item Mild sleep apnea: 5 $\leq$ AHI $\leq$ 15
    \item Moderate sleep apnea: 15 $\leq$ AHI $\leq$ 30
    \item Severe sleep apnea: AHI $\geq$ 30
\end{itemize}
\smallskip

\begin{figure*}[!t]
\centerline{\includegraphics[width=\textwidth]{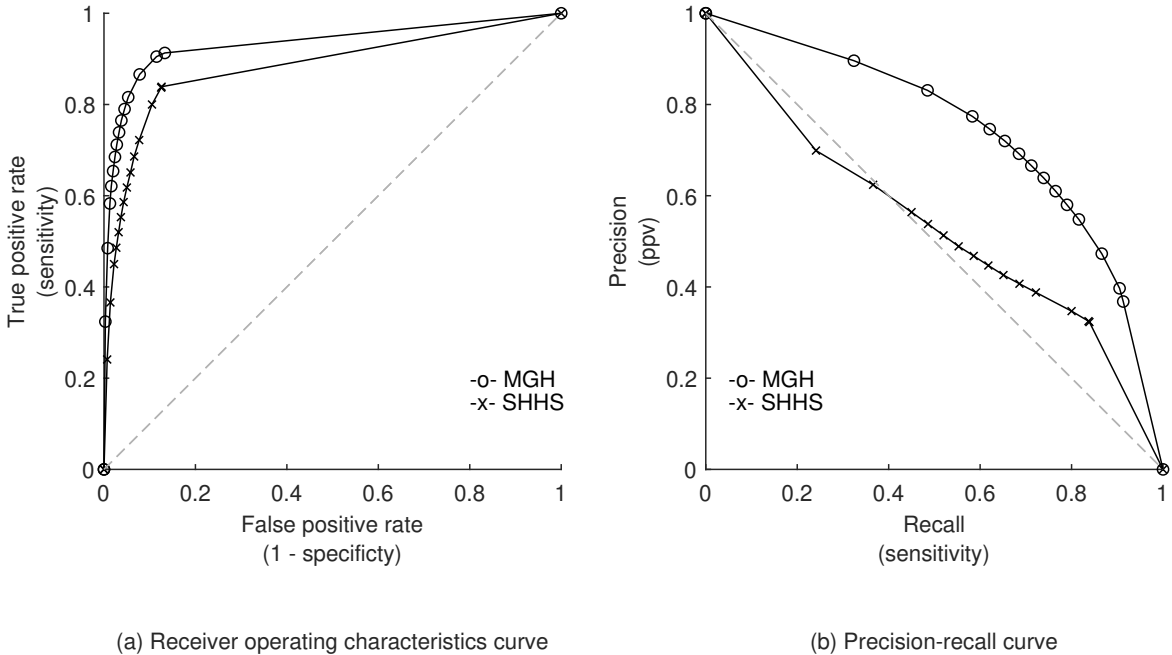}}
\centering
\caption{ROC and PRC curves for binary classification in experiment 1.}
\label{fig2}
\end{figure*}

We obtained the classification accuracy of our model by creating a confusion matrix for the four AHI scores. The classification accuracy displays the ability of the model to assign a patient to any of the four AHI categories. To gain insight into accuracy of the AHI prediction disregarding the discrete borders used in categorization, histograms were computed to show the difference between the AHI value scored by the experts and the AHI value predicted by our model. For both experiments, scatter plots visualizing the correlation between the expert-scored AHI and the algorithm-predicted AHI were computed. Additionally, for experiment 2, we computed scatter plots for the RDI and each type of respiratory event per hour of sleep. A robust linear regression model with bi-squared cost function was fitted to the data to compute the correlation between the scored AHI by the experts and predicted AHI by our model \cite{fitRobustMatlab}. This model was selected to mitigate the effect of outliers.

\newpage
\section{Results}
For the MGH and SHHS testing dataset, 16 and 129 recordings, respectively, were removed due to insufficient sleep or erroneous data. The boosted model approach resulted in 5 and 8 consecutive model iterations before reaching the desired class balance in experiment 1 and 2, respectively. The median length among all respiratory events was 18 seconds. This length was used to determine the number of true negative epochs.

\subsection{Per-event performance}
The binary model showed accuracy values of 95\% for the MGH dataset and 94\% for the SHHS dataset, resulting from specificity values of 97\% and 99\%, and sensitivity values of 71\% and 66\% for the two datasets respectively. An AUC\textsubscript{ROC} value of 0.93 and AUC\textsubscript{PRC} of 0.74 were found for the MGH dataset, and AUC\textsubscript{ROC} and AUC\textsubscript{PRC} for the SHHS dataset were 0.88 and 0.53 (see Fig. \ref{fig2} for the ROC and PRC curves). An overview of all per-event performance metrics for the binary task are given in Table \ref{tab3}. Both absolute and normalized confusion matrices for MGH and SHHS dataset are shown in Appendix Table \ref{app:tab1}.

\begin{table}
\centering
\caption{Overall per-event performance for experiment 1 and 2 with all values in percentages}
\small
\begin{tabular}{|l|c c c|}
    \hline
     & \multicolumn{2}{c}{Experiment 1} & Experiment 2 \\ 
     &&& \\
     & MGH dataset & SHHS dataset & MGH dataset \\
     & \footnotesize(binary) & \footnotesize(binary) & \footnotesize(multiclass) \\ 
     \hline
    Accuracy       & 95 & 94 & 96 \\
    Sensitivity    & 71 & 66 & 43 \\
    Specificity     & 97 & 95 & 99 \\
    Precision      & 68 & 39 & 41 \\
    F1-score       & 70 & 49 & 39 \\
    AUC\textsubscript{ROC} & 93 & 88 & - \\
    AUC\textsubscript{PRC} & 74 & 53 & - \\
    \hline
\end{tabular}
\label{tab3}
\end{table}

In experiment 2, the multiclass model resulted in an overall accuracy of 96\%. Mean performance metrics over the four respiratory event classes are 43\% sensitivity, 99\% specificity, 41\% precision and 39\% F1-score. Performances vary considerably for the different classes, e.g. while 81\% of all expert-labeled central apnea events are correctly classified, this is only true for 16\% of hypopneas. See Table \ref{tab4} for the performance metrics for all classes. The absolute and normalized confusion matrices are shown in Appendix Table \ref{app:tab1}.

\begin{table}
\centering
\caption{Per-event performance in experiment 2. All values are percentages}
\small
\begin{tabular}{|l|c c c c|}
    \hline
    \small Experiment 2 & Sensitivity & Specificity & Precision & F1-score \\
    \hline
    Obstructive apnea  & 46 & 99 & 45 & 45\\
    Central apnea      & 81 & 99 & 50 & 62\\
    RERA               & 29 & 97 & 26 & 27\\
    Hypopnea           & 16 & 99 & 42 & 23\\
    \hline
    Mean               & 43 & 99 & 41 & 39 \\
    \hline
\end{tabular}
\label{tab4}    
\end{table}

\subsection{Per-patient performance}
We next assessed the performance of our model to classify the severity of AHI. For the MGH dataset in experiment 1, performance among each AHI subgroup is shown in Table \ref{tab5}. The sensitivity, precision and F1-score increased with the severity of apnea. The opposite effect was observed for the accuracy and specificity.

\begin{table*}
\centering
\caption{Per-event performance per AHI subgroup for the MGH dataset in experiment 1 with all values in percentages}
\begin{tabular}{|r|c c c c c|}
    \hline
     & Accuracy & Sensitivity & Specificity & Precision & F1-score \\
    \hline
    \small{Normal breathing \scriptsize{(N=360)}}   & 93 & 42 & 99 & 40 & 41\\
    \small{Mild apnea \scriptsize{(N=268)}}         & 94 & 56 & 97 & 56 & 56\\
    \small{Moderate apnea \scriptsize{(N=227)}}     & 93 & 70 & 95 & 71 & 71\\
    \small{Severe apnea \scriptsize{(N=129)}}       & 90 & 79 & 94 & 83 & 81\\
    \hline
\end{tabular}
\label{tab5}    
\end{table*}

\begin{figure*}[!t]
\centering
\centerline{\includegraphics[width=\textwidth]{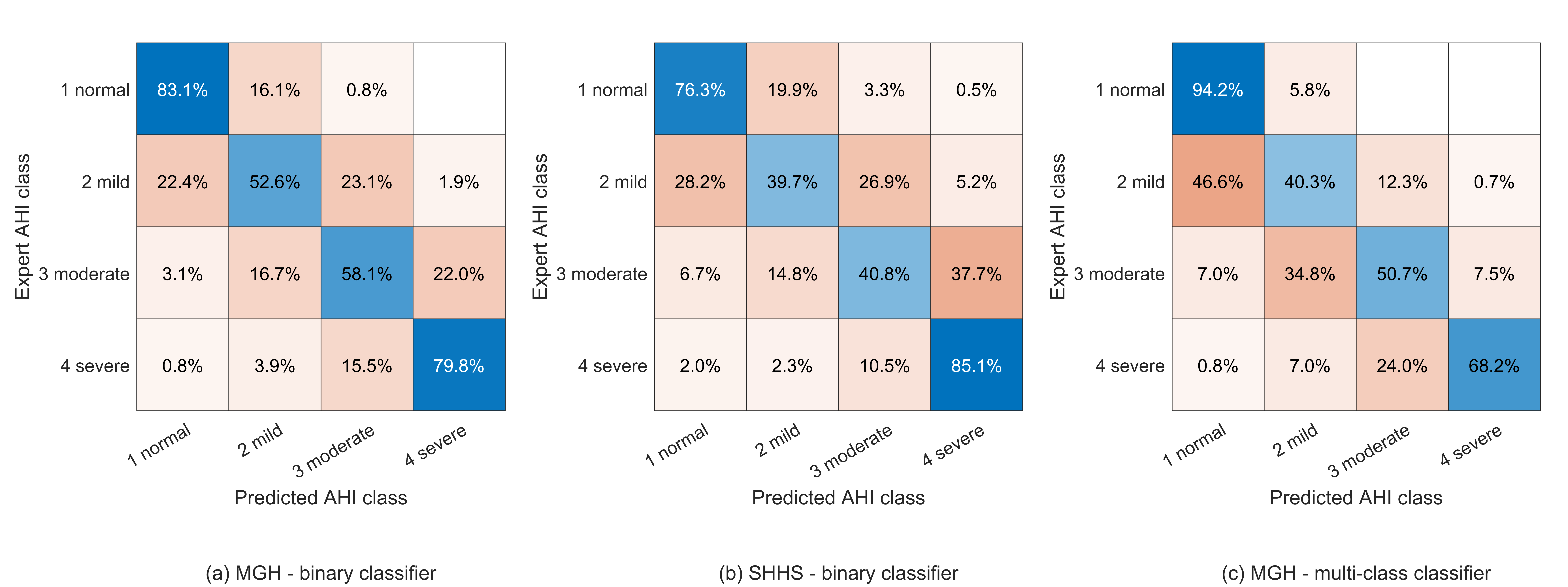}}
\caption{AHI classification confusion matrices for both experiments.}
\label{fig3}
\end{figure*}

We computed the confusion matrix for AHI prediction, as shown in Fig. \ref{fig3}. Overall, 69\% of patients from the MGH dataset and 58\% from the SHHS dataset were assigned to the correct AHI category. In experiment 2, 65\% of all patients were classified in the correct AHI category using the MGH dataset. As expected, misclassifications resulted in false positives in the neighboring AHI categories. 

\begin{figure*}
\centering
\centerline{\includegraphics[width=\textwidth]{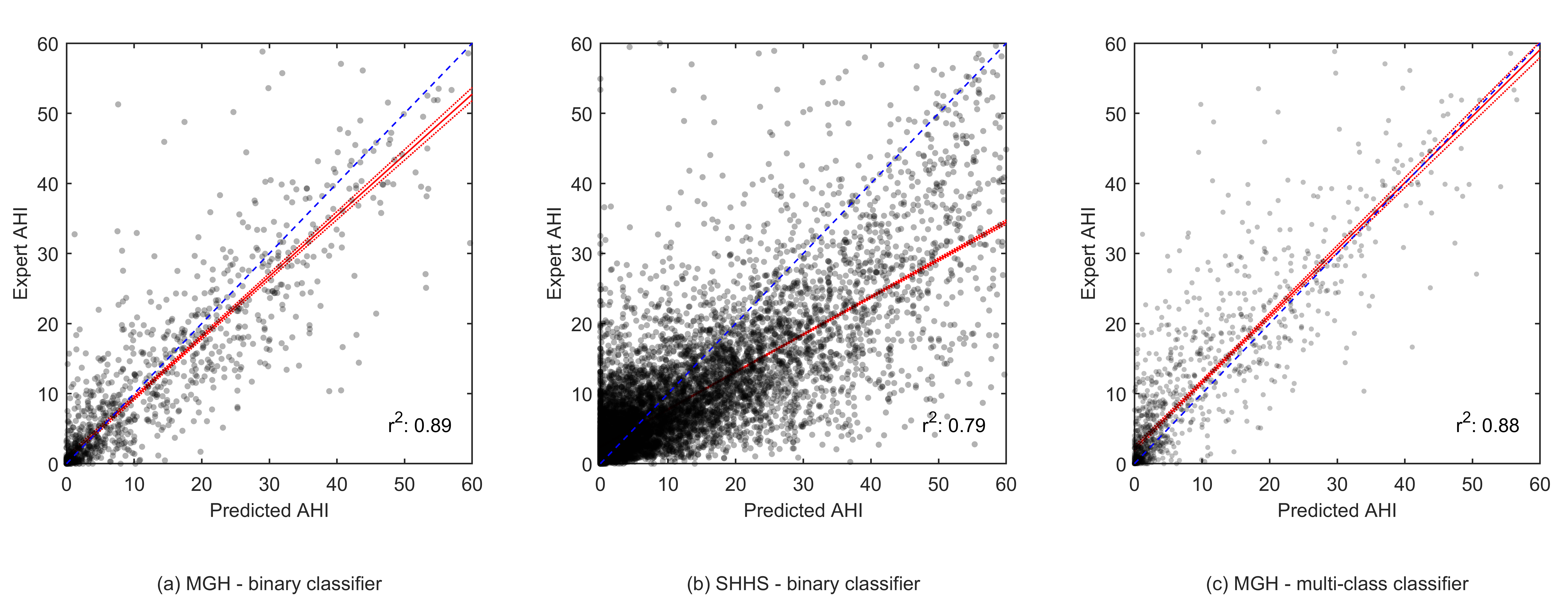}}
\caption{Scatter plots showing the correlation between the expert-scored AHI and the model predicted AHI in experiment 1 and 2. The fitted robust linear regression model is shown in red.}
\label{fig4}
\end{figure*}

The scatter plots in Fig. \ref{fig4} show the correlation between the expert-scored AHI and the model predicted AHI from experiment 1. The r\textsuperscript{2} was 0.89 for the MGH dataset and 0.79 for the SHHS dataset. For experiment 2, an r\textsuperscript{2} of 0.88, 0.82, 0.90, 0.98, 0.32 and 0.58 was determined for AHI, RDI, obstructive apneas, central apneas, RERA’s and hypopneas, respectively, see Appendix Fig. \ref{app:fig1}. 

\begin{figure*}[!t]
\centering
\centerline{\includegraphics[width=\textwidth]{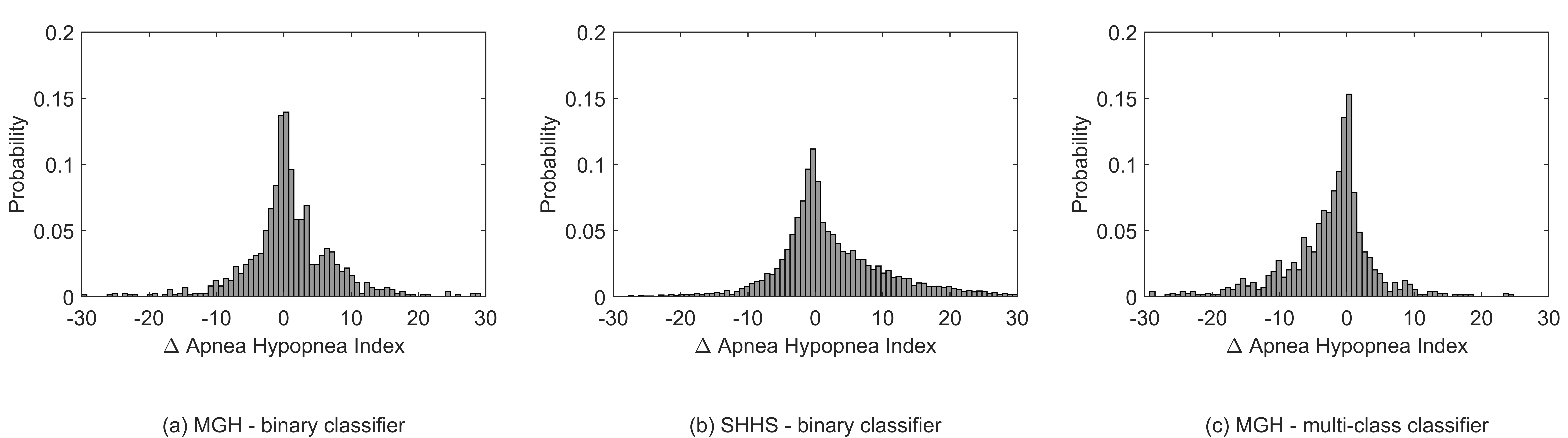}}
\caption{Histograms of the relative difference in AHI determined by the experts vs the model in experiment 1 and 2.}
\label{fig5}
\end{figure*}

The computed histograms represent the difference between the AHI value scored by the experts and the AHI value predicted by our model. The error distributions showed a clear peak around 0 with a standard deviation of 7.7 AHI for the MGH dataset in both experiments and a standard deviation of 9.7 for the SHHS dataset, see Fig. \ref{fig5}.

\section{Discussion}
A deep neural network method was developed to classify typical breathing disorders during sleep based on a single respiratory effort belt used in PSG. In a first experiment our WaveNet model successfully discriminated respiratory events from regular breathing on our primary dataset with an accuracy of 95\%, and sensitivity, specificity, precision and F1-score of 71\%, 97\%, 68\% and 70\%, respectively. AHI was predicted for each patient from the number of respiratory events. An accuracy of 69\% was found for the AHI category prediction task. It is notable that most misclassifications of our model resulted in false positives into the neighboring AHI categories. This effect is best visualized in the histograms in Fig. \ref{fig5}; the unimodal and symmetrical shape shows that a decrease in number of false positives was observed as the difference between the predicted AHI and the sleep-expert scored AHI increased. The correlation between expert-scored AHI and algorithm-predicted AHI showed an r\textsuperscript{2} of 0.89. 

When applying our model on a secondary dataset obtained from the SHHS, a slight decrease in model performance was observed. This is likely due to imperfect generalization to a dataset where different respiratory effort sensors are used. It is important to note that a thermistor was used to detect respiratory events in the SHHS study, whereas a nasal pressure sensor was used in the MGH study. The nasal pressure transducer is more sensitive in detecting sleep-disordered breathing than the thermistor \cite{Budhiraja2005}. Therefore, it is likely that a significant number of events were missed during annotation in the SHHS study. When observing the performance of our model applied on the SHHS dataset a decrease of approximately 20\% was observed for the precision and f1-score. Sensitivity, however, was only reduced by 5\%. This observation can be explained by the different methodology used while annotating events. The fact that our model generally overpredicts AHI when applied on the SHHS dataset, as seen in the scatter plot of Fig. \ref{fig4} (b), is in line with this assumption. 

The guidelines for scoring respiratory events manually have evolved over the years but have remained largely driven by consensus. Thus, for example, the requirement of a 50\% or 30\% reduction in signal amplitude is arbitrary; there is no data to suggest that a 35\% or 60\% would be less or more clinically meaningful. Moreover, visual discrimination of small percentage differences is likely poor. During polysomnography or even home sleep study recordings, the multichannel nature of the data enables increased scoring accuracy by associating changes with neighboring signals. Moreover, airway collapse is common during central apnea, and high loop gain can drive obstructive events. Thus, the differentiation of obstructive and central events is not as pathophysiologically clear as clinical scoring may suggest. This biological reality of blurred boundaries will be reflected in any manual or automated scoring approach. 

In a secondary experiment our model successfully identified the type of the included respiratory events, i.e. central apneas, obstructive apneas, hypopneas and RERA’s. Despite a similar overall accuracy, discrimination of the specific respiratory events resulted in a decreased per-event performance with respect to the first experiment. Central apneas were detected with high sensitivity of 81\%, expectedly due to the apparent effect of the disorder on respiratory effort. Often markedly reduced respiratory effort is observed during central apnea events, resulting in clear features for algorithms to recognize. We expect that this is the main reason of the high-performance metrics for the detection of central apneas. This is true to a lesser extent for obstructive apnea events, hence the slightly lower performance when compared to the central apneas. The recognition of hypopneas and RERA’s was considered poor, with an F1-score of 23\% and 27\% respectively. The scatter plots show underprediction by our model, indicating limited sensitivity rather than specificity. Without additional information derived from other physiological signals the identification of hypopneas and RERA’s appears difficult. It should be noted that scoring RERAs and central hypopneas are considered so difficult that the AASM scoring guidelines leaves these as “optional”, and most clinical services do not score such events. There are also several biological inconsistencies with the conventional rules for scoring central hypopneas, adding to the probability of misclassification during “gold standard” scoring. 

However, misclassification during multiclassification often meant that an event of a particular respiratory class was classified as a different class. When observing per-patient performance of our multiclassification model, large variation in performance was observed among the various respiratory events. Yet, when the different predicted classes are grouped together to binary apnea events, a similar correlation was found between the expert-scored AHI and the algorithm-predicted AHI. An r\textsuperscript{2} of 0.88 was determined, indicating that AHI prediction based on the specific respiratory events is feasible. An overall slightly reduced performance was observed in AHI prediction confusion matrices with respect to the binary classification of experiment 1. The ability to discriminate various respiratory events is clinically valuable but may not be achievable by using manual scoring as a gold standard. The type of breathing assistance and overall apnea treatment may vary for different underlying pathology leading to apnea. Specifying the type of apnea will therefore provide aid in improving personalized patient care. 

Besides a high accuracy, a metric that is affected by class imbalance, our model also showed high AUC values for ROC (0.93) and PRC (0.71) which means the model not only has an excellent agreement in sensitivity and specificity but also has a clinically acceptable precision.

Most approaches found in the literature used different sensors to detect respiratory events. Some have shown slightly higher performances, although performance comparisons are difficult given the different datasets and evaluation methods. To our knowledge, our model showed better results with respect to other methods using a single respiratory effort belt and is the only model that shows that additional respiratory event class discrimination is possible.

An advantage of using an effort belt to assess apnea is the non-invasive application. This becomes very relevant when assessing respiratory stability and instability/events in intensive care or environmentally hostile conditions. Using limited resources – such as a respiratory effort belt – to assess respiratory abnormalities can be successfully applied in combination with other simple and small sensors necessary for monitoring patients in diverse clinical situations. Patients receiving breathing aid using CPAP are eligible for event detection. The number of patients included in our research is larger than previous reports in the literature. This, in combination with limited preprocessing and without the use of any human-engineered features, emphasizes the robustness of our proposed approach. 

\section{Conclusion}
A neural network approach to analyzing typical respiratory events during sleep based on a single respiratory measurement is described. Our model included dilated convolutions to allow their receptive fields to grow exponentially with depth, which is important to model the long-range temporal dependencies in respiration signals. Using this model, we obtained a comparable performance with respect to literature while using a minimally invasive methodology. The use of a respiratory effort belt at the abdomen for sleep apnea analysis bears the advantage of wide implementation options ranging from acute care settings to wearable devices for home usage. Important first steps were obtained in automated apnea detection with limited resources, creating new sleep assessment opportunities applicable to the clinical setting.

\appendices

\renewcommand*{\UrlFont}{\rmfamily}
\renewcommand*{\bibfont}{\footnotesize}
\printbibliography

\section{Additional Tables and Figures}

\begin{table*}
\centering
\caption{Confusion matrices for experiments 1 and 2 in both absolute and relative values.}
\small
    \begin{tabular}{c}
        \begin{tabular}{l r}
            \begin{tabular}{|l||c c|}
                \hline
                Experiment 1, MGH       & predicted    & predicted \\
                absolute values         & No-event     & event \\
                \hline
                \hline
                True, No-event      & \cellcolor{Gray}971480    & 28133 \\
                True, event         & 23166 & \cellcolor{Gray}57121     \\
                \hline
            \end{tabular}
             & 
            \begin{tabular}{|l||c c|}
                \hline
                Experiment 1, SHHS      & predicted    & predicted \\
                absolute values         & No-event     & event \\
                \hline
                \hline
                True, No-event      & \cellcolor{Gray}8025098    & 410268 \\
                True, event         & 134018 & \cellcolor{Gray}262781     \\
                \hline
            \end{tabular}
        \end{tabular} \\ \\ 
        \begin{tabular}{l r}
            \begin{tabular}{|l||c c|}
                \hline
                Experiment 1, MGH       & predicted     & predicted \\
                normalized values       & No-event      & event \\
                \hline
                \hline
                True, No-event      & \cellcolor{Gray}0.97      & 0.03      \\
                True, event         & 0.29          & \cellcolor{Gray}0.71  \\ 
                \hline
            \end{tabular}
             & 
             \begin{tabular}{|l||c c|}
                \hline
                Experiment 1, SHHS      & predicted     & predicted \\
                normalized values       & No-event      & event \\
                \hline
                \hline
                True, No-event      & \cellcolor{Gray}0.95      & 0.05      \\
                True, event         & 0.34          & \cellcolor{Gray}0.66  \\ 
                \hline
            \end{tabular}
        \end{tabular}

   \\ \\ \hline \hline \\
   
   \begin{tabular}{|l||c c c c c|}
            \hline
            Experiment 2, MGH       & predicted     & predicted & predicted & predicted & predicted \\
            absolute values         & No-event      & Obstructive & Central & RERA      & Hypopnea  \\
            \hline
            \hline
            True, No-event      & \cellcolor{Gray}970484    & 2234      & 5427      & 20032     & 3975  \\
            True, Obstructive   & 4394          & \cellcolor{Gray}10998 & 3339      & 2781      & 2619  \\
            True, Central       & 1791          & 1150      & \cellcolor{Gray}13924 & 191       & 124    \\
            True, RERA          & 21907         & 1990      & 1854      & \cellcolor{Gray}11367 & 2665  \\
            True, Hypopnea      & 14883         & 7982      & 3333      & 9520      & \cellcolor{Gray}6893  \\
            \hline
        \end{tabular}
 \\ \\
        \begin{tabular}{|l||c c c c c|}
            \hline
            Experiment 2, MGH       & predicted     & predicted & predicted & predicted & predicted \\
            normalized values       & No-event      & Obstructive & Central & RERA      & Hypopnea \\
            \hline
            \hline
            True, No-event      & \cellcolor{Gray}0.97      & 0.0       & 0.01      & 0.02      & 0.0    \\
            True, Obstructive   & 0.18          & \cellcolor{Gray}0.46  & 0.14      & 0.12      & 0.11  \\  
            True, Central       & 0.1           & 0.07      & \cellcolor{Gray}0.81  & 0.01      & 0.01  \\
            True, RERA          & 0.55          & 0.05      & 0.05      & \cellcolor{Gray}0.29  & 0.07  \\
            True, Hypopnea      & 0.35          & 0.19      & 0.08      & 0.22      & \cellcolor{Gray}0.16  \\
            \hline
        \end{tabular}
        \\ \\
        
    \end{tabular}
    \label{app:tab1}
\end{table*}

\begin{figure*}
\centering
\centerline{\includegraphics[width=\textwidth]{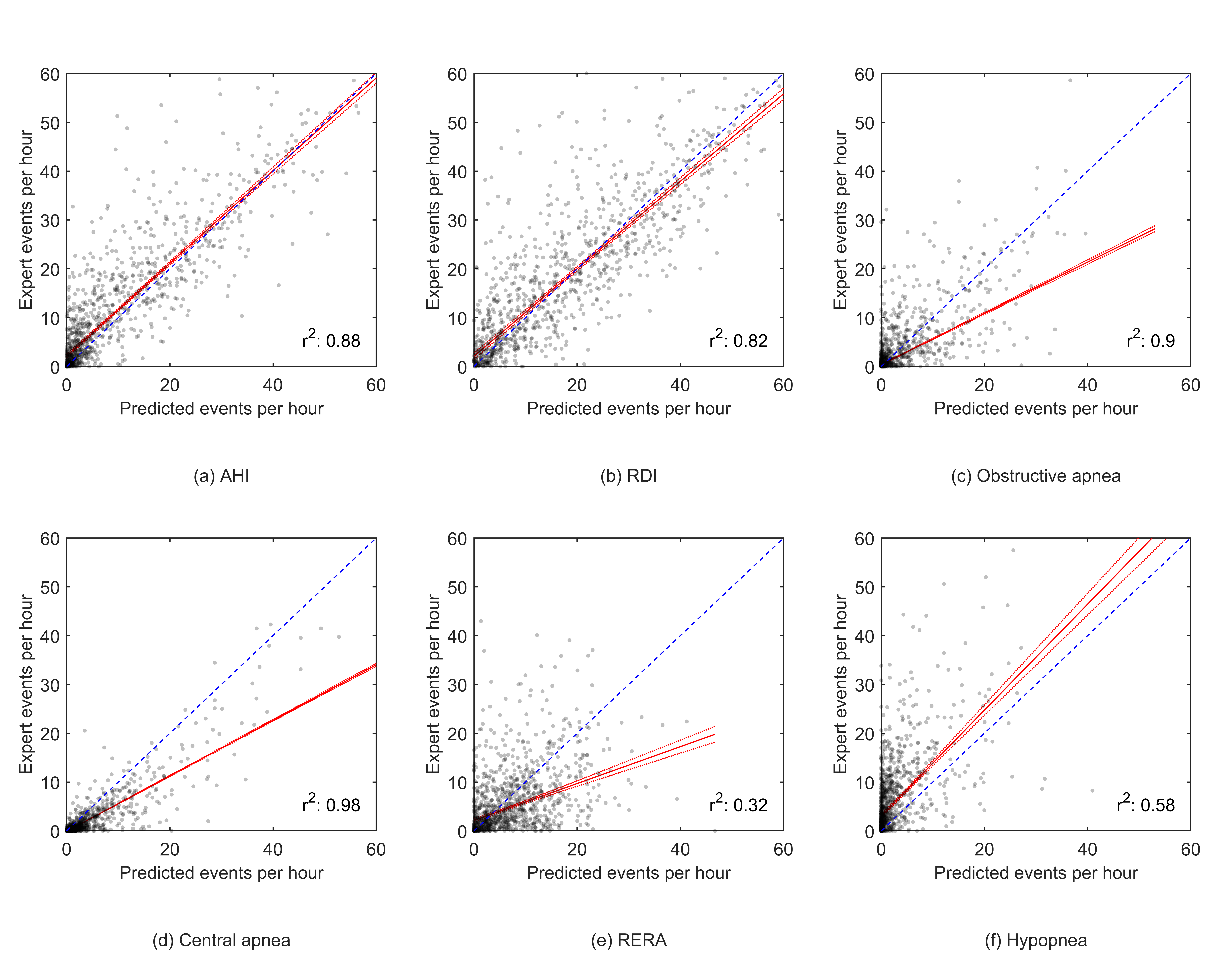}}
\caption{Scatter plots showing the correlation between the expert-scored respiratory events and the model predicted respiratory events from experiment 2. The fitted robust linear regression model is shown in red.}
\label{app:fig1}
\end{figure*}

\end{document}